\documentclass[a4paper,twoside]{article}

\usepackage{times}
\usepackage{bm}
\usepackage{tabularx}
\usepackage{epsfig}
\usepackage{subcaption}
\usepackage{calc}
\usepackage{amssymb}
\usepackage{amstext}
\usepackage{siunitx}
\usepackage{booktabs}
\usepackage{algorithmic}
\usepackage[table,xcdraw]{xcolor}
\usepackage{amsmath}
\usepackage{graphicx}
\usepackage{amsthm, amsfonts}
\usepackage{multicol}
\usepackage{multirow}
\usepackage{textcomp}
\usepackage{pslatex}
\usepackage{float}
\usepackage[normalem]{ulem}
\useunder{\uline}{\ul}{}
\newcommand{\etal}{et al.}
\usepackage{arydshln}
\usepackage[space]{cite}
\usepackage[normalem]{ulem}
\newcommand{\minus}{\scalebox{0.5}[1.0]{$-$}}
\usepackage[pagebackref=true,breaklinks=true,colorlinks,bookmarks=false]{hyperref}
\def\BibTeX{{\rm B\kern-.05em{\sc i\kern-.025em b}\kern-.08em
    T\kern-.1667em\lower.7ex\hbox{E}\kern-.125emX}}

\usepackage{apalike}
\usepackage[section]{placeins}
\usepackage{SCITEPRESS}     

\makeatletter
\newcommand{\printfnsymbol}[1]{%
  \textsuperscript{\@fnsymbol{#1}}%
}

\begin{document}

\title{Transformers in Self-Supervised Monocular Depth Estimation with Unknown Camera Intrinsics}

\author{\authorname{Arnav Varma\thanks{Equal contribution}, Hemang Chawla\textcolor{red}{\footnotemark[1]}, Bahram Zonooz and Elahe Arani}
\affiliation{Advanced Research Lab, NavInfo Europe, The Netherlands}
\email{\{arnav.varma, hemang.chawla, elahe.arani\}@navinfo.eu, bahram.zonooz@gmail.com}
}



\abstract{The advent of autonomous driving and advanced driver assistance systems necessitates continuous developments in computer vision for 3D scene understanding. Self-supervised monocular depth estimation, a method for pixel-wise distance estimation of objects from a single camera without the use of ground truth labels, is an important task in 3D scene understanding. However, existing methods for this task are limited to convolutional neural network (CNN) architectures. 
In contrast with CNNs that use localized linear operations and lose feature resolution across the layers, vision transformers process at constant resolution with a global receptive field at every stage. 
While recent works have compared transformers against their CNN counterparts for tasks such as image classification, no study exists that investigates the impact of using transformers for self-supervised monocular depth estimation.
Here, we first demonstrate how to adapt vision transformers for self-supervised monocular depth estimation.
Thereafter, we compare the transformer and CNN-based architectures for their performance on KITTI depth prediction benchmarks, as well as their robustness to natural corruptions and adversarial attacks, including when the camera intrinsics are unknown. Our study demonstrates how transformer-based architecture, though lower in run-time efficiency, achieves comparable performance while being more robust and generalizable.}

\onecolumn \maketitle \normalsize \setcounter{footnote}{0} \vfill

\section{\uppercase{Introduction}}
\label{sec:introduction}
\begin{figure*}[htbp]
\centering
  \includegraphics[width=\linewidth]{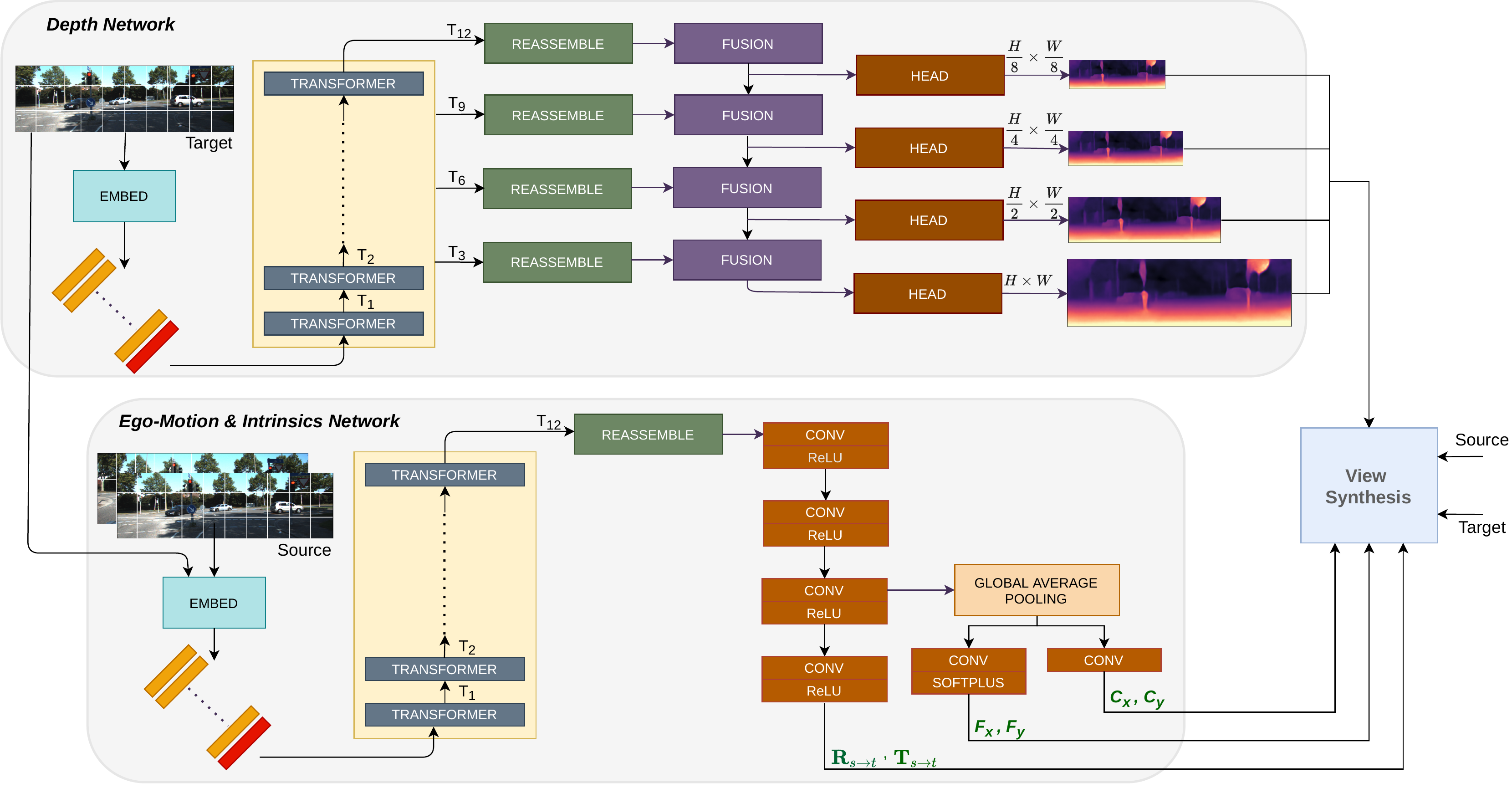}
  \caption{An overview of 
  Monocular Transformer Structure from Motion Learner (MT-SfMLearner) with learned intrinsics. We readapt modules from  Dense Prediction Transformer (DPT) and Monodepth2 to be trained with appearance-based losses for self-supervised monocular depth, ego-motion, and intrinsics estimation.} 
\label{fig:architecture}
\end{figure*}




There have been rapid improvements in scene understanding for robotics and advanced driver assistance systems (ADAS) over the past years. This success is attributed to the use of Convolutional Neural Networks (CNNs) within a mostly encoder-decoder paradigm. Convolutions provide spatial locality and translation invariance which has proved useful for image analysis tasks. The encoder, often a convolutional Residual Network (ResNet)~\cite{he2016deep}, learns feature representations from the input and is followed by a decoder which aggregates these features and converts them into final predictions. However, the choice of architecture has a major impact on the performance and generalizability of the task.


While CNNs have been the preferred architecture in computer vision, transformers have also recently gained traction~\cite{dosovitskiy2020vit} motivated by their success in natural language processing~\cite{vaswani2017attention}. Notably, they have also outperformed CNNs for object detection~\cite{carion2020end} and semantic segmentation~\cite{zheng2021rethinking}. This is also reflected in methods for monocular dense depth estimation, a pertinent task for autonomous planning and navigation, where supervised transformer-based methods~\cite{li2020revisiting,Ranftl2021} have been proposed as an alternative to supervised CNN-based methods~\cite{lee2019big,aich2020bidirectional}. However, supervised methods require extensive RGB-D ground truth collected from costly LiDARs or multi-camera rigs. Instead, self-supervised methods have increasingly utilized concepts of Structure from Motion (SfM) with known camera intrinsics to train monocular depth and ego-motion estimation networks simultaneously~\cite{guizilini20203d,lyu2020hr,chawlavarma2021multimodal}. While transformer ingredients such as attention have been utilized for self-supervised depth estimation~\cite{johnston2020self}, most methods are nevertheless limited to the use of CNNs that have localized linear operations and lose feature resolution during downsampling to increase their limited receptive field~\cite{yang2021transformers}. 

On the other hand, transformers with fewer inductive biases allow for more globally coherent predictions with different layers attending to local and global features simultaneously~\cite{touvron2021training}.
However, transformers require more training data and can be more computationally demanding~\cite{caron2021emerging}. 
While multiple studies have compared transformers against CNNs for tasks such as image classification~\cite{raghu2021vision,bhojanapalli2021understanding}, no study exists that evaluates the impact of transformers in self-supervised monocular depth estimation, including when the camera intrinsics may be unknown.


In this work, we conduct a comparative study between CNN- and transformer-based architectures for self-supervised monocular depth estimation. Our contributions are as follows:
\begin{itemize}
    \item We demonstrate how to adapt vision transformers 
    for self-supervised monocular depth estimation by implementing a method called Monocular-Transformer SfMLearner (MT-SfMLearner). 
    \item We compare MT-SfMLearner and CNNs for their performance  on the KITTI monocular depth Eigen Zhou split~\cite{eigen2014depth} and the online depth prediction benchmark~\cite{geiger2013vision}. 
    \item We investigate the impact of architecture choices for the individual depth and ego-motion networks on performance as well as robustness to natural corruptions and adversarial attacks. 
    \item We also introduce a modular method that simultaneously predicts camera focal lengths and principal point from the images themselves and can easily be utilized within both CNN- and transformer-based architectures. 
    \item We study the accuracy of intrinsics estimation as well as its impact on the performance and robustness of depth estimation.  
    \item Finally, we also compare the run-time computational and energy efficiency of the architectures for depth and intrinsics estimation. 
\end{itemize}

MT-SfMLearner provides real-time depth estimates and illustrates how transformer-based architecture, though lower in run-time efficiency, can achieve comparable performance as the CNN-based architectures while being more robust under natural corruptions and adversarial attacks, even when the camera intrinsics are unknown. 
Thus, our work presents a way to analyze the trade-off between the performance, robustness, and efficiency of transformer- and CNN-based architectures for depth estimation. 

\section{\uppercase{Related Works}}
\label{sec:related}
Recently, transformer architectures such as Vision Transformer (ViT)~\cite{Ranftl2021} and  Data-efficient image Transformer (DeiT)~\cite{touvron2021training} have outperformed CNN architectures in image classification. Studies comparing ViT and CNN architectures like ResNet have further demonstrated that transformers are more robust to natural corruptions and adversarial examples in classification~\cite{bhojanapalli2021understanding,paul2021vision}.
Motivated by their success, researchers have replaced CNN encoders with transformers in scene understanding tasks such as object detection~\cite{carion2020end,liu2021swin}, semantic segmentation~\cite{zheng2021rethinking,strudel2021segmenter}, and supervised monocular depth estimation~\cite{Ranftl2020,yang2021transformers}. 

For \textit{supervised} monocular depth estimation, Dense Prediction Transformer (DPT)~\cite{Ranftl2020} uses ViT as the encoder with a convolutional decoder and shows more coherent predictions than CNNs due to the global receptive field of transformers.
TransDepth~\cite{yang2021transformers} additionally uses a ResNet projection layer and attention gates in the decoder to induce the spatial locality of CNNs for \textit{supervised} monocular depth and surface-normal estimation. 
Lately, some works have inculcated elements of transformers such as self-attention~\cite{vaswani2017attention} in \textit{self-supervised} monocular depth estimation~\cite{johnston2020self,xiang2021self}.
However, there has been no investigation of transformers to replace the traditional CNN-based methods~\cite{godard2019digging,lyu2020hr} for \textit{self-supervised} monocular depth estimation.

Moreover, self-supervised monocular depth estimation still requires prior knowledge of the camera intrinsics (focal length and principal point) during training, which may be different for each data source, may change over time, or be unknown a priori~\cite{chawla2020crowdsourced}. While multiple approaches to \textit{supervised} camera intrinsics estimation have been proposed~\cite{lopez2019deep,zhuang2019degeneracy}, not many \textit{self-supervised} approaches exist~\cite{gordon2019depth}.

Therefore, we investigate the impacts of transformer architectures on self-supervised monocular depth estimation for their performance, robustness, and run-time efficiency, even when intrinsics are unknown.
\section{\uppercase{Method}}
\label{sec:method}

Our objective is to study the effect of utilizing vision transformers for self-supervised monocular depth estimation in contrast with the contemporary methods that utilize Convolutional Neural Networks. 

Given a set of $n$ images from a video sequence, we simultaneously train depth and ego-motion prediction networks. The inputs to the networks are a sequence of temporally consecutive RGB image triplets $\{I_{\minus 1}, I_0, I_1\} \in \mathbb{R}^{H\times W \times 3}$. The depth network learns the model $f_D:\mathbb{R}^{H\times W \times 3} \rightarrow \mathbb{R}^{H \times W}$ to output dense depth (or disparity) for each pixel coordinate $p$ of a single image. Simultaneously, the ego-motion network learns the model $f_{E}:\mathbb{R}^{2 \times H\times W \times 3} \rightarrow \mathbb{R}^6$ to output relative translation $(t_x, t_y, t_z)$ and rotation $(r_x, r_y, r_z)$  forming the affine transformation  \tiny $\begin{bmatrix} \hat{{R}} & \hat{{T}} \\ {0} & 1 \end{bmatrix}$ \normalsize $\in \text{SE(3)}$ between a pair of overlapping images.  
The predicted depth $\hat{{D}}$ and ego-motion $\hat{{T}}$ are linked together via the perspective projection model,
\begin{equation}
\label{eq:perspective_model}
    p_s \sim K\hat{{R}}_{s \leftarrow t} \hat{{D}}_t(p_t)K^{-1}p_t + K\hat{{T}}_{s\leftarrow t},
\end{equation}
that warps the source images $I_s \in \{I_{\minus 1}, I_1\}$ to the target image $I_t \in \{I_0\}$, with the camera intrinsics described by $K$. This process is called view synthesis, as shown in Figure~\ref{fig:architecture}. We train the networks using the appearance-based \textit{photometric} loss between the real and synthesized target images, as well as a \textit{smoothness} loss on the depth predictions~\cite{godard2019digging}.

\begin{table*}[!h]
    \centering
    \caption{Architecture details of the \textit{Reassemble} modules. DN and EN refer to depth and ego-motion networks, respectively. The subscripts of $DN$ refer to the transformer layer from which the respective \textit{Reassemble} module takes its input (see Figure \ref{fig:architecture}). Input image size is $H \times W$, $p$ refers to the patch size, $N_p=H\cdot W / p^2$ refers to the number of patches from the image, and $d$ refers to the feature dimension of the transformer features.}
\resizebox{\linewidth}{!}{
\begin{tabular}{|l|c|c|c|c|}
\hline
\textbf{Operation}             & \textbf{Input size}                      & \textbf{Output size}                       & \textbf{Function}                                                          & \begin{tabular}[c]{@{}c@{}}\textbf{Parameters}\\ ($DN_3$, $DN_6$, $DN_9$, $DN_{12}$, $EN$)\end{tabular} \\ \hline \hline
Read                  & $(N_p + 1) \times d$          & $N_p \times d$                  & Drop readout token                                                & $-$                                                                            \\
Concatenate           & $N_p \times d$                & $d \times H / p \times W / p$   & Transpose and Unflatten                                           & $-$                                                                            \\
Pointwise Convolution & $d \times H / p \times W / p$ & $N_c \times H / p \times W / p$   & $N_c$ channels                                                    & $N_c=[96, 768, 1536, 3072, 2048]$                                              \\
Strided Convolution   & $N_c \times H / p \times W / p$ & $N_c \times H / 2p \times W / 2p$ & $k \times k$ convolution, Stride$=2$, $N_c$ channels, padding$=1$ & $k=[-, -, -, 3, -]$                                                            \\
Transpose Convolution & $N_c \times H / p \times W / p$ & $N_c \times H / s \times W / s$   & $p/s \times p/s$ deconvolution, stride$=p/s$, $N_c$ channels      & $s=[4, 2, -, -, -]$                                                            \\ \hline
\end{tabular}
}
    \label{tab:reassemble}
\end{table*}

\begin{table}[b]
    \centering
    \caption{Architecture details of \textit{Head} modules in Figure \ref{fig:architecture}.}
\resizebox{0.83\linewidth}{!}{\begin{tabular}{|c|}
\hline
\textbf{Layers}                                                                            \\ \hline \hline
 $32$ $3 \times 3$ \textit{Convolutions}, stride=$1$, padding$=1$ \\ \hdashline
\textit{ReLU}                                                    \\ \hdashline
\textit{Bilinear Interpolation} to upsample by $2$                \\ \hdashline
$32$ \textit{Pointwise Convolutions}                                                   \\ \hdashline
\textit{Sigmoid}                                                 \\ \hline
\end{tabular}}
    \label{tab:heads}
\end{table}

\subsection{Architecture}\label{subsec:architecture}
Here we describe Monocular Transformer Structure from Motion Learner (MT-SfMLearner), our transformer-based method for self-supervised monocular depth estimation (Figure~\ref{fig:architecture}).

\subsubsection{Depth Network}
For the depth network, we readapt the Dense Prediction Transformer (DPT)~\cite{Ranftl2020} for \textit{self-supervised learning}, with a DeiT-Base~\cite{touvron2021training} in the encoder.
There are five components of the depth network:
\begin{itemize}
    \item An \textbf{\textit{Embed}} module, which is a part of the encoder, takes an image $I \in \mathbb{R}^{H\times W \times 3}$, and converts non-overlapping image patches of size $p \times p$ into $N_{p} = H\cdot W / p^2$ tokens $t_i \in \mathbb{R}^{d} \ \forall i \in [1, 2,... N_p]$, where $d=768$ for DeiT-Base. This is implemented as a large $p \times p$ convolution with stride $s = p$ where $p = 16$. The output from this module is concatenated with a \textit{readout} token of the same size as the remaining tokens.
    \item The \textbf{\textit{Transformer}} block, that is also a part of the encoder, consists of $12$ transformer layers which process these tokens with multi-head self-attention (MHSA)~\cite{vaswani2017attention} modules. MHSA processes inputs at constant resolution and can simultaneously attend to global and local features.
    \item Four \textbf{\textit{Reassemble}} modules in the decoder, which are responsible for extracting image-like features from the $3$\textsuperscript{rd}, $6$\textsuperscript{th}, $9$\textsuperscript{th}, and $12$\textsuperscript{th} (final) transformer layers by dropping the readout token and concatenating the remaining tokens in $2$D. This is followed by pointwise convolutions to change the number of channels, and transpose convolution in the first two reassemble  modules to upsample the representations (corresponding to $T_3$ and $T_6$ in Figure \ref{fig:architecture}). 
    To make the transformer network comparable to its convolutional counterparts, we increase the number of channels in the pointwise convolutions of the last three \textit{Reassemble} modules by a factor of 4 with respect to 
    DPT. The exact architecture of the \textit{Reassemble} modules can be found in Table \ref{tab:reassemble}.
    \item Four \textbf{\textit{Fusion}} modules in the decoder, based on RefineNet~\cite{lin2017refinenet}. They progressively fuse information from the \textit{Reassemble} modules with information passing through the decoder, and upsample the features by $2$ at each stage. Unlike DPT, we enable batch normalization in the decoder as it was found to be helpful for self-supervised depth prediction. 
    We also reduce the number of channels in the \textit{Fusion} block to $96$ from $256$ in 
    DPT. 
    \item Four \textbf{\textit{Head}} modules at the end of each \textit{Fusion} module to predict depth at $4$ scales following previous self-supervised methods \cite{godard2019digging}. 
    Unlike DPT, the \textit{Head} modules use $2$ convolutions instead of $3$ as we found no difference in performance.
    For further details of the \textit{Head} modules, refer to Table \ref{tab:heads}.
\end{itemize}

\subsubsection{Ego-Motion Network}
For the ego-motion network, we adapt DeiT-Base \cite{touvron2021training} in the encoder. Since the input to the transformer for the ego-motion network consists of two images concatenated along the channel dimension, we repeat the embedding layer accordingly. We use a \textit{Reassemble} module to pass transformer tokens to the decoder. For details on the structure of this \textit{Reassemble} module, refer to Table~\ref{tab:reassemble}.
We adopt the decoder for the ego-motion network from Monodepth2~\cite{godard2019digging}.

When both depth and ego-motion networks use transformers as described above, we refer to the resulting architecture as Monocular Transformer Structure from Motion Learner (\textit{MT-SfMLearner}).


\subsection{Appearance-based Losses}
Following contemporary self-supervised monocular depth estimation methods, we adopt the \textit{appearance-based losses} and an \textit{auto-masking} procedure from the CNN-based Monodepth2~\cite{godard2019digging} for the above described transformer-based architecture as well. 
We employ a photometric reprojection loss composed of the pixel-wise $\ell_1$ distance and the Structural Similarity (SSIM) between the real and synthesized target images, along with a multi-scale edge-aware \textit{smoothness} loss on the depth predictions. We also use auto-masking to disregard the temporally stationary pixels in the image triplets. Furthermore, we reduce texture-copy artifacts by calculating the total loss after upscaling the depths, predicted at 4 scales, from intermediate decoder layers to the input resolution.

\subsection{Intrinsics}\label{intrinsics}
Accurate camera intrinsics given by
\begin{equation}
\label{eq:camera_matrix}
    K = \begin{bmatrix}
    f_x & 0 & c_x\\
    0 & f_y & c_y\\
    0 & 0 & 1
    \end{bmatrix},
\end{equation}
are essential to self-supervised depth estimation as can be seem from Equation \ref{eq:perspective_model}.
However, the intrinsics may vary within a dataset with videos collected from different camera setups, or over a long period of time.  These parameters can also be unknown for crowdsourced datasets.  

We address this by introducing an intrinsics estimation module. We modify the ego-motion network, which takes a pair of consecutive images as input, and learns to estimate the focal length and principal point along with the translation and rotation.
Specifically, we add a convolutional path in the ego-motion decoder to learn the intrinsics. The decoder features before activation from the penultimate layer are passed through a global average pooling layer, followed by two branches of pointwise convolutions to reduce the number of channels from $256$ to $2$. One branch uses a softplus activation to estimate focal lengths along $x$ and $y$ axes as the focal length is always positive. The other branch doesn't use any activation to estimate the principal point as it has no such constraint.  Note that the ego-motion decoder is the same for both the convolutional as well as transformer architectures. Consequently, the intrinsics estimation method can be modularly utilized with both architectures.  Figure \ref{fig:architecture}  demonstrates MT-SfMLearner with learned intrinsics. 

\section{\uppercase{Results}}
\label{sec:results}
In this section, we perform a comparative analysis between our transformer-based method, MT-SfMLearner, and the existing CNN-based methods for self-supervised monocular depth estimation. We also perform a contrastive study on the architectures for the depth and ego-motion networks to evaluate their impact on the prediction accuracy, robustness to natural corruptions and adversarial attacks, and the run-time computational and energy efficiency.
Finally, we analyze the correctness and run-time efficiency of intrinsics predictions, and also study its impact on the accuracy and robustness of depth estimation. 


\subsection{Implementation Details}
\label{subsec:details}
\subsubsection{Dataset } 
We report all results on the Eigen Split~\cite{eigen2014depth} of KITTI~\cite{geiger2013vision} dataset after removing the static frames as per \cite{zhou2017unsupervised}, unless stated otherwise. This split contains $39,810$ training, $4424$ validation, and $697$ test images, respectively. All results are reported on the per-image scaled dense depth prediction without post-processing~\cite{godard2019digging}, unless stated otherwise. 

\subsubsection{Training Settings}


The networks are implemented in PyTorch~\cite{paszke2019pytorch} and trained on a TeslaV100 GPU for 20 epochs at a resolution of $640\times192$ with batch-size 12.
MT-SfMLearner is further trained at 2 more resolutions - $416\times128$ and $1024\times320$, with batch-sizes of 12 and 2, respectively for experiments in Section~\ref{subsec:depth_table}. 
The depth and ego-motion encoders are initialized with ImageNet~\cite{deng2009imagenet} pre-trained weights. 
We use the Adam~\cite{kingma2014adam} optimizer for CNN-based networks (in Sections \ref{subsec:ablation} and \ref{subsec:intrinsics}) and AdamW~\cite{loshchilov2017decoupled} optimizer for transformer-based networks with initial learning rates of $1e^{-4}$ and $1e^{-5}$, respectively. 
The learning rate is decayed after 15 epochs by a factor of 10. Both optimizers use $\beta_{1}=0.9$ and $\beta_{2}=0.999$. 

\begin{table*}[tb]
\centering
\caption{Quantitative results comparing MT-SfMLearner with existing methods on KITTI Eigen split. For each category of image sizes, the best results are displayed in bold, and the second best results are underlined.}
\resizebox{\textwidth}{!}{
\begin{tabular}{|c|l|c|cccc|ccc|}
\hline
\multirow{2}{*}{}  &\multirow{2}{*}{\textbf{Methods}} & \multirow{2}{*}{\textbf{Resolution}} & \multicolumn{4}{c|}{\textbf{\cellcolor{red!25}Error$\downarrow$}} & \multicolumn{3}{c|}{\textbf{\cellcolor{blue!25}Accuracy$\uparrow$}} \\ \cline{4-10}
 &    &  & Abs Rel & Sq Rel & RMSE & RMSE log & $\delta<1.25$ & $\delta<1.25^2$ & $\delta<1.25^3$ \\ \hline \hline
\multirow{8}{*}{\rotatebox[origin=c]{90}{LR}}
 & SfMLearner\cite{zhou2017unsupervised} & 416$\times$128 & 0.208 & 1.768 & 6.856 & 0.283 & 0.678 & 0.885 & 0.957 \\
  & GeoNet~\cite{yin2018geonet} & 416$\times$128 & 0.155 & 1.296 & 5.857 & 0.233 & 0.793 & 0.931 & 0.973 \\
  & Vid2Depth~\cite{mahjourian2018unsupervised} & 416$\times$128 & 0.163 & 1.240 & 6.220 & 0.250 & 0.762 & 0.916 & 0.968 \\
  & Struct2Depth~\cite{casser2019depth} & 416$\times$128 & 0.141 & 1.026 & 5.291 & 0.215 & 0.816 & 0.945 & 0.979 \\
  & Roussel \etal~\cite{roussel2019monocular} & 416$\times$128 & 0.179 & 1.545 & 6.765 & 0.268 & 0754 & 0916 & 0.966 \\
  & VITW~\cite{gordon2019depth} & 416$\times$128 & 0.129 & \underline{0.982} & 5.230 & 0.213 & 0.840 & 0.945 & 0.976 \\
  & Monodepth2 
  ~\cite{godard2019digging} & 416$\times$128 & \underline{0.128} & 1.087 & \underline{5.171} & \underline{0.204} & \textbf{0.855} & \textbf{0.953} & \underline{0.978} \\
  & \textbf{MT-SfMLearner (Ours)} & 416$\times$128 & \textbf{0.125} & \textbf{0.905} & \textbf{5.096} & \textbf{0.203} & \underline{0.851} & \underline{0.952} & \textbf{0.980} \\
\hline
  
  \multirow{10}{*}{\rotatebox[origin=c]{90}{MR}}  
  & CC~\cite{ranjan2019competitive} & 832$\times$256 & 0.140 & 1.070 & 5.326 & 0.217 & 0.826 & 0.941 & 0.975 \\
  & SC-SfMLearner~\cite{bian2019unsupervised} & 832$\times$256 & 0.137 & 1.089 & 5.439 & 0.217 & 0.830 & 0.942 & 0.975 \\
  & Monodepth2 
  ~\cite{godard2019digging} & 640$\times$192 & 0.115 & 0.903 & 4.863 & 0.193 & 0.877 & 0.959 & 0.981 \\
  & SG Depth~\cite{klingner2020selfsupervised} & 640$\times$192 & 0.117 & 0.907 & 4.844 & 0.194 & 0.875 & 0.958 & 0.980 \\
  & PackNet-SfM~\cite{guizilini20203d} & 640$\times$192 & 0.111 & \underline{0.829} & 4.788 & 0.199 & 0.864 & 0.954 & 0.980 \\
  & Poggi et. al~\cite{poggi2020uncertainty} & 640$\times$192 & 0.111 & 0.863 & 4.756 & \underline{0.188} & 0.881 & \underline{0.961} & \underline{0.982} \\
  & Johnston \& Carneiro~\cite{johnston2020self} & 640$\times$192 & \textbf{0.106} & 0.861 & \underline{4.699} & \textbf{0.185} & \textbf{0.889} & \textbf{0.962} & \underline{0.982} \\
  & HR-Depth~\cite{lyu2020hr} & 640$\times$192 & \underline{0.109} & \textbf{0.792} & \textbf{4.632} & \textbf{0.185} & \underline{0.884} & \textbf{0.962} & \textbf{0.983} \\  
  & \textbf{MT-SfMLearner (Ours)} & 640$\times$192 & 0.112 & 0.838 & 4.771 & \underline{0.188} & 0.879 & 0.960 & \underline{0.982} \\
  \hline
  
\multirow{4}{*}{\rotatebox[origin=c]{90}{HR}}  
 & Packnet-SfM~\cite{guizilini20203d} & 1280$\times$384 & 0.107 & 0.803 & 4.566 & 0.197 & 0.876 & 0.957 & 0.979 \\
 & HR-Depth~\cite{lyu2020hr} & 1024$\times$320 & \underline{0.106} & \textbf{0.755} & \textbf{4.472} & \textbf{0.181} & \underline{0.892} & \textbf{0.966} & \textbf{0.984} \\  
 & G2S~\cite{chawlavarma2021multimodal} & 1024$\times$384 & 0.109 & 0.844 & 4.774 & \underline{0.194} & 0.869 & 0.958 & 0.981 \\
 & \textbf{MT-SfMLearner (Ours)} & 1024$\times$320 & \textbf{0.104} & \underline{0.799} & \underline{4.547} & \textbf{0.181} & \textbf{0.893} & \underline{0.963} & \underline{0.982} \\\hline
\end{tabular}}

\label{tab:depth_table}
\end{table*}

\subsection{Depth Estimation Performance}
\label{subsec:depth_table}
First we compare MT-SfMLearner, where both depth and ego-motion networks are transformer-based, with the existing fully convolutional neural networks for their accuracy on self-supervised monocular depth estimation.
Their performance is evaluated using metrics from~\cite{eigen2014depth} up to a fixed range of \SI{80}{m} 
as shown in Table \ref{tab:depth_table}.  We compare the methods at different input image sizes clustered into categories of Low-Resolution (LR), Medium-Resolution (MR), and High-Resolution (HR). We do not compare against methods that use ground-truth semantic labels during training.
All methods assume known ground-truth camera intrinsics. 

We observe that MT-SfMLearner 
is able to achieve comparable performance at all resolutions under \textit{error} as well as \textit{accuracy} metrics. This includes methods that also utilize a heavy encoder such as ResNet-101 ~\cite{johnston2020self} and PackNet~\cite{guizilini20203d}. 


\paragraph{Online Benchmark:}
We also measure the performance of MT-SfMLearner on the KITTI Online Benchmark for depth prediction\footnote{ \url{http://www.cvlibs.net/datasets/kitti/eval_depth.php?benchmark=depth_prediction}. See under MT-SfMLearner.}  using the metrics from~\cite{uhrig2017sparsity}. We train on an image size of $1024 \times 320$, and add the G2S loss~\cite{chawlavarma2021multimodal} for obtaining predictions at metric scale.  Results ordered by their rank are shown in Table \ref{tab:kitti_depth_online}.
The performance of MT-SfMLearner is on par with state-of-the-art self-supervised methods, and outperforms several supervised methods. This further confirms that the transformer-based method can achieve comparable performance to the convolutional neural networks for self-supervised depth estimation.

\begin{table}[b]
\centering
\caption{Quantitative comparison of \textit{unscaled} dense depth prediction on the KITTI Depth Prediction Benchmark (online server). Supervised training with ground truth
depths is denoted by D.  Use of monocular sequences or stereo pairs is represented by M and S, respectively. Seg represents additional supervised semantic segmentation training. The use of GPS for multi-modal self-supervision is denoted by G.}
\resizebox{\columnwidth}{!}{
\begin{tabular}{|l|c|c|c|c|c|}
\hline 
\textbf{Method}                 & \textbf{Train}            & \cellcolor{red!25}\textbf{SILog$\downarrow$} & \cellcolor{red!25}\textbf{SqErrRel$\downarrow$}  \\ \hline \hline
DORN       ~\cite{fu2018deep}                   & D & 11.77 & 2.23  \\
SORD       ~\cite{diaz2019soft}                 & D & 12.39 & 2.49  \\
VNL        ~\cite{yin2019enforcing}             & D & 12.65 & 2.46  \\
DS-SIDENet ~\cite{ren2019deep}                  & D & 12.86 & 2x.87  \\

PAP        ~\cite{zhang2019pattern}             & D & 13.08 & 2.72  \\
Guo \etal  ~\cite{guo2018learning}              & D+S & 13.41 & 2.86  \\
G2S        ~\cite{chawlavarma2021multimodal}    & M+G & 14.16 & 3.65  \\
\hline \hline
Ours                                            & M+G & 14.25 & 3.72  \\
\hline \hline
Monodepth2 ~\cite{godard2019digging}            & M+S & 14.41 & 3.67  \\
DABC       ~\cite{li2018deep}                   & D & 14.49 & 4.08  \\
SDNet      ~\cite{ochs2019sdnet}                & D+Seg & 14.68 & 3.90  \\
APMoE      ~\cite{kong2019pixel}                & D & 14.74 & 3.88  \\
CSWS       ~\cite{li2018monocular}              & D & 14.85 & 348  \\

HBC        ~\cite{jiang2019hierarchical}        & D & 15.18 & 3.79  \\
SGDepth    ~\cite{klingner2020selfsupervised}   & M+Seg & 15.30 & 5.00 \\
DHGRL      ~\cite{zhang2018deep}                & D & 15.47 & 4.04 \\
PackNet-SfM ~\cite{guizilini20203d}             & M+V & 15.80 & 4.73 \\
MultiDepth ~\cite{liebel2019multidepth}         & D & 16.05 & 3.89 \\
LSIM       ~\cite{goldman2019learn}             & S & 17.92 & 6.88 \\
Monodepth  ~\cite{godard2017unsupervised}       & S & 22.02 & 20.58 \\ 
\hline
\end{tabular}}
\label{tab:kitti_depth_online}
\end{table}

\subsection{Contrastive Study}
\label{subsec:ablation}
We saw in the previous section that MT-SfMLearner performs competently on independent and identically distributed (i.i.d.) test set with respect to the state-of-the-art. However, networks that perform well on an i.i.d. test set may still learn shortcut features that are non-robust and generalize poorly to out-of-distribution (o.o.d.) datasets~\cite{geirhos2020shortcut}. Since self-supervised monocular depth estimation networks concurrently train an ego-motion network (see Equation \ref{eq:perspective_model}), we investigate the impact of each network's architecture on depth estimation. 

We consider both Convolutional (C) and Transformer (T) networks for depth and ego-motion estimation. The resulting four combinations of (Depth Network, Ego-Motion Network) architectures are (C, C), (C, T), (T, C), and (T, T), ordered on the basis of their increasing influence of transformers on depth estimation. 
To compare our transformer-based method fairly with convolutional networks, we utilize Monodepth2~\cite{godard2019digging} with a ResNet-101~\cite{he2016deep} encoder. All four combinations are trained thrice using the settings described in Section \ref{subsec:details} for an image size of $640 \times 192$. All combinations assume known ground-truth camera intrinsics. 
\subsubsection{Performance}
\label{ablation-performance}

\begin{table*}[htbp]
\centering
\caption{Quantitative results on KITTI Eigen split for all four architecture combinations of depth and ego-motion networks. The best results are displayed in bold, the second best are underlined.} 
\resizebox{0.8\textwidth}{!}{
\begin{tabular}{|c|cccc|ccc|}
\hline
\multirow{2}{*}{\textbf{Architecture}}  & \multicolumn{4}{c|}{\textbf{\cellcolor{red!25}Error$\downarrow$}} & \multicolumn{3}{c|}{\textbf{\cellcolor{blue!25}Accuracy$\uparrow$}} \\ \cline{2-8}
       & Abs Rel & Sq Rel & RMSE & RMSE log & $\delta<1.25$ & $\delta<1.25^2$ & $\delta<1.25^3$ \\ \hline \hline
  C, C & \textbf{0.111} & 0.897 & 4.865 & 0.193 & \textbf{0.881} & \underline{0.959} & 0.980 \\ \hdashline
  C, T & 0.113 & 0.874 & 4.813 & 0.192 & \underline{0.880} & \textbf{0.960} & \underline{0.981} \\
  T, C  & \underline{0.112} & \underline{0.843} & \textbf{4.766} & \underline{0.189} & 0.879 & \textbf{0.960} & \textbf{0.982} \\ \hdashline
  T, T  & 0.112 & \textbf{0.838} & \underline{4.771} & \textbf{0.188} & 0.879 & \textbf{0.960} & \textbf{0.982} \\
  \hline
\end{tabular}
}
\label{tab:ablation_table}
\end{table*}

\begin{figure*}[t]
\centering
  \includegraphics[width=\linewidth]{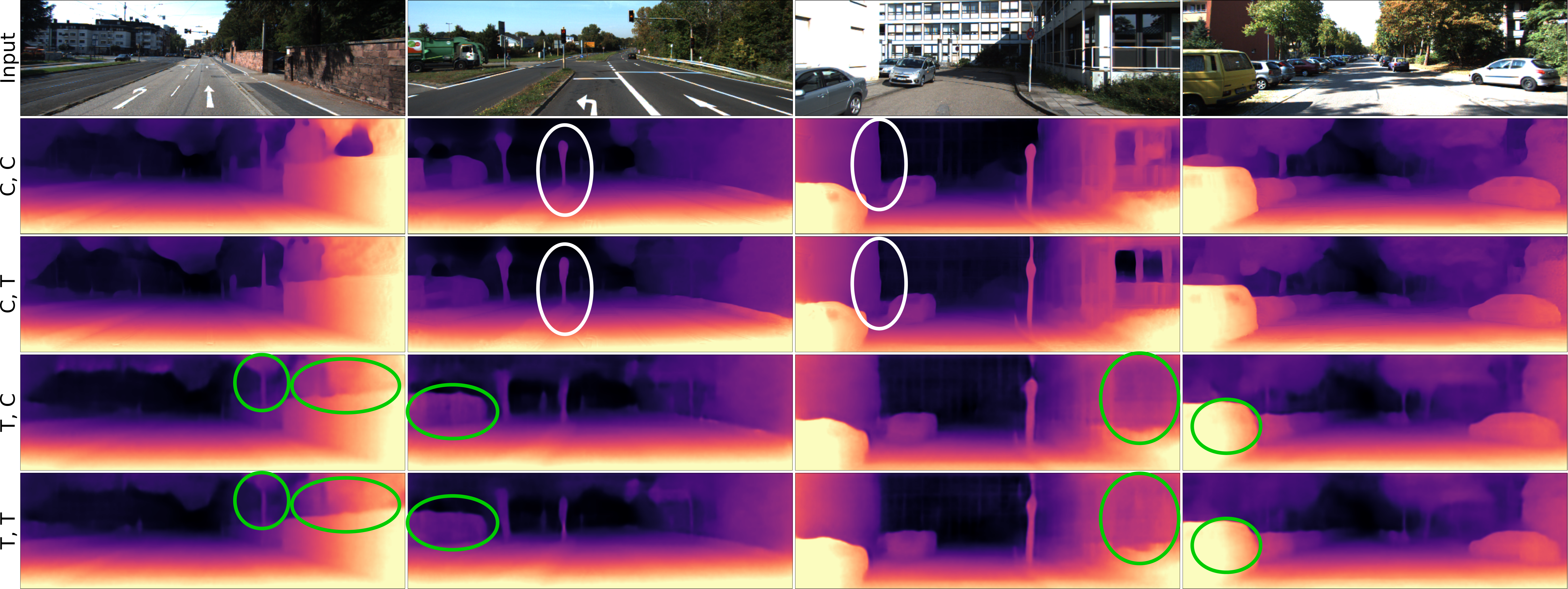}
  \caption{Disparity maps on KITTI for qualitative comparison of all four architecture combinations of depth and ego-motion networks. Example areas where the global receptive field of transformer is advantageous are highlighted in green. Example areas where local receptive field of CNNs is advantageous are highlighted in white.
  } 
\label{fig:ablation_vis}
\end{figure*}

For the four combinations, we report the best performance on i.i.d. in Table \ref{tab:ablation_table}, and visualize the depth predictions for the same in Figure \ref{fig:ablation_vis}.  The i.i.d. test set used for comparison is same as in Section \ref{subsec:depth_table}.

We observe from Table \ref{tab:ablation_table} that 
the combination of transformer-based depth and ego-motion networks i.e MT-SfMLearner performs best under two of the \textit{error} metrics as well as two of the \textit{accuracy} metrics. The remaining combinations perform comparably on all the metrics.

From Figure \ref{fig:ablation_vis},  we observe more uniform estimates for larger objects like vehicles, vegetation, and buildings when depth is learned using transformers.
Transformers are also less affected by reflection from windows of vehicles and buildings. 
This is likely because of the larger receptive fields of the self-attention layers, which lead to more globally coherent predictions. On the other hand, convolutional networks produce sharper boundaries, 
and perform better on thinner objects
such as traffic signs and poles. 
This is likely because of the inherent inductive bias for spatial locality present in convolutional layers.

\subsubsection{Robustness}
\label{ablation-robustness}
While the different combinations perform comparably on the i.i.d. dataset, they may differ in robustness and generalizability. 
Therefore, we study the impact of natural corruptions and adversarial attacks on the depth performance using the following:
\begin{itemize}
    \item \textbf{Natural corruptions}. Following~\cite{hendrycks2019robustness} and ~\cite{michaelis2019dragon}, we generate 15 corrupted versions of the KITTI i.i.d. test set at the highest severity($=5$). These natural corruptions fall under 4 categories - \textit{noise} (Gaussian, shot, impulse), \textit{blur} (defocus, glass, motion, zoom), \textit{weather} (snow, frost, fog, brightness), and \textit{digital} (contrast, elastic, pixelate, JPEG). 
    \item \textbf{Projected Gradient Descent (PGD) adversarial examples}.  Adversarial attacks make imperceptible (to humans) changes to input images to create adversarial examples that fool networks. 
    We generate adversarial examples from the i.i.d. test set using PGD~\cite{DBLP:conf/iclr/MadryMSTV18} at attack strength $\epsilon \in \{0.25, 0.5, 1.0, 2.0, 4.0, 8.0, 16.0, 32.0\}$. The gradients are calculated with respect to the training loss. Following ~\cite{kurakin2016adversarial}, the perturbation is accumulated over $min(\epsilon + 4, \lceil1.25\cdot\epsilon\rceil)$ iterations with a step-size of $1$. When the test image is from the beginning or end of a KITTI sequence, the appearance-based loss is only calculated for the feasible pair of images. 
    \item \textbf{Symmetrically flipped adversarial examples}. Inspired by ~\cite{wong2020targeted}, we generate these adversarial examples to fool the networks into predicting flipped estimates. For this, we use the gradients of the RMSE loss, where the targets are symmetrical horizontal and vertical flips of the i.i.d. predictions. This evaluation is conducted at attack strength $\epsilon \in \{1.0, 2.0, 4.0\}$, similar to the PGD attack described above.
\end{itemize}

\begin{figure*}[h]
\centering
\includegraphics[width=0.83\linewidth]{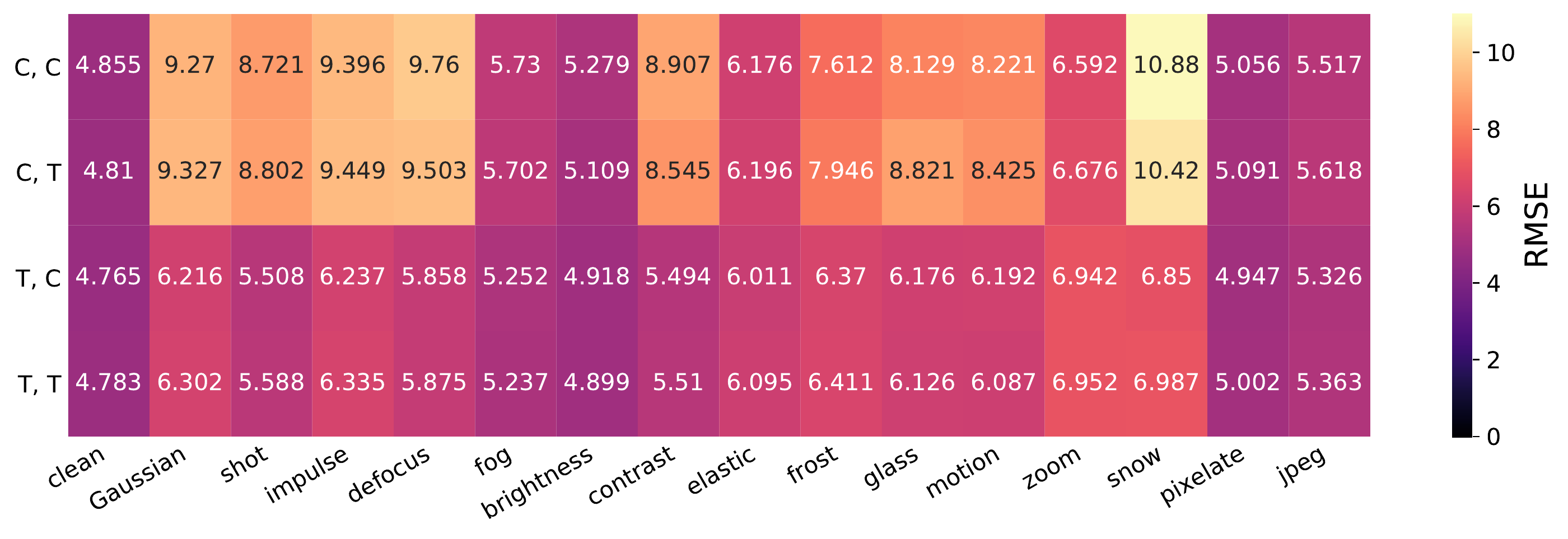}
\caption{RMSE for natural corruptions of KITTI for all four combinations of depth and ego-motion networks. The i.i.d. evaluation is denoted by \textit{clean}.
}
\label{fig:heatmap_corruption}
\end{figure*}

\begin{figure}[h]
\centering
\includegraphics[width=\linewidth]{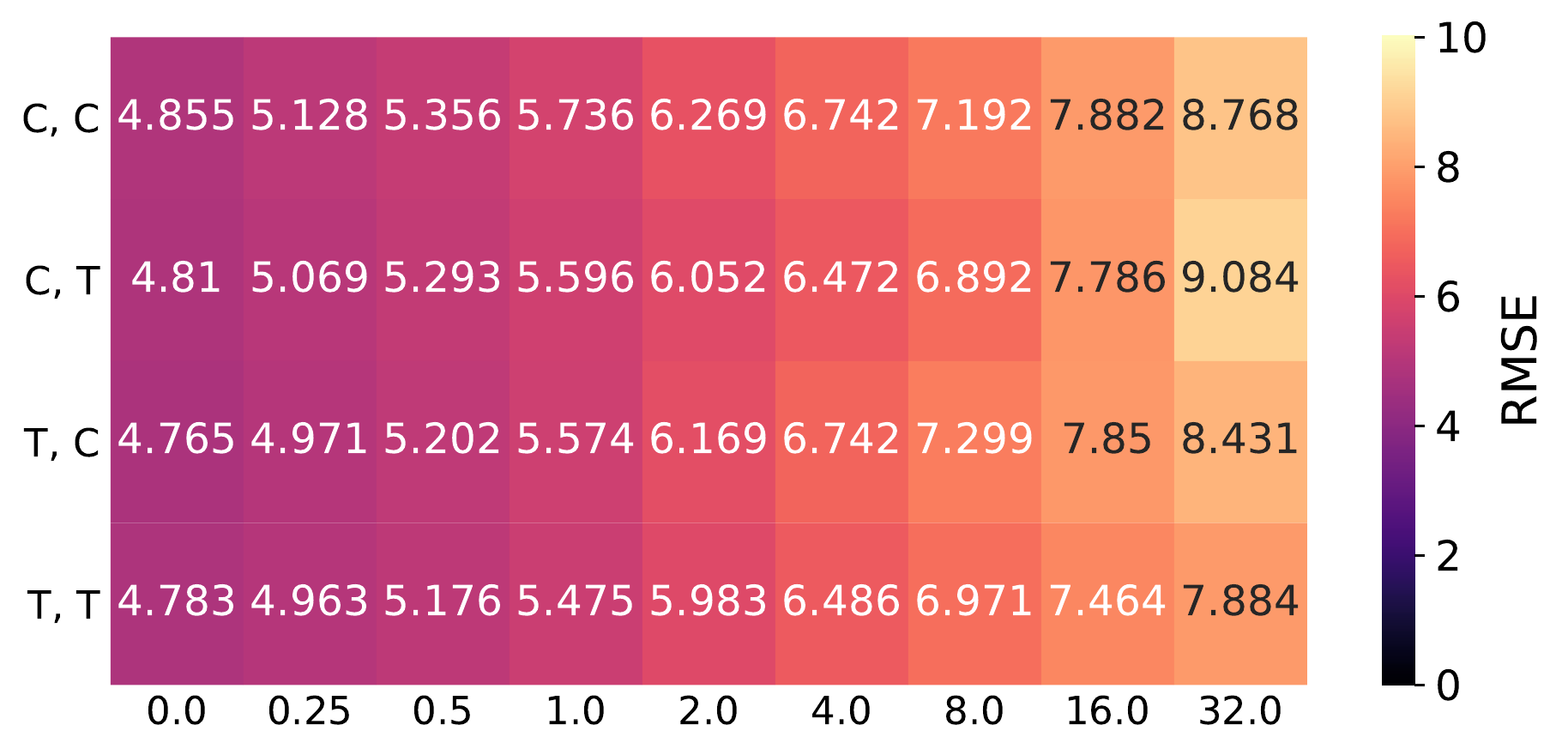}
\caption{RMSE for adversarial corruptions of KITTI generated using PGD at all attack strengths (0.0 to 32.0) for all four combinations of depth and ego-motion networks. Attack strength $0.0$ refers to i.i.d. evaluation.
}
\label{fig:heatmap_pgd}
\end{figure}

\begin{figure}[h]
\centering
\includegraphics[width=\linewidth]{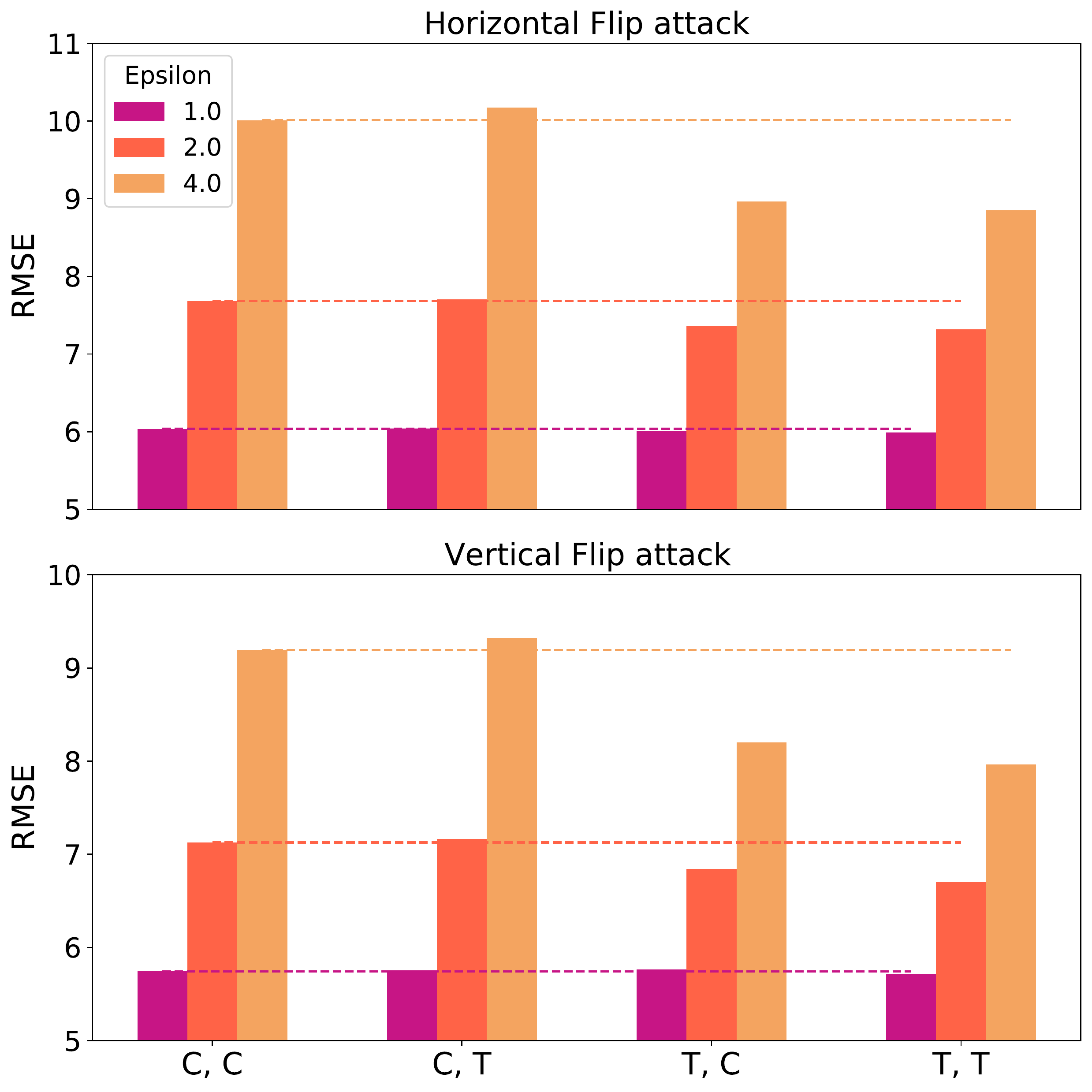}
\caption{RMSE for adversarial corruptions of KITTI generated using horizontal and vertical flip attacks for all four combinations of depth and ego-motion networks.
}
\label{fig:bar_flip}
\end{figure}


We report the mean RMSE across three training runs on natural corruptions, PGD adversarial examples, and symmetrically flipped adversarial examples in Figures \ref{fig:heatmap_corruption}, \ref{fig:heatmap_pgd}, and \ref{fig:bar_flip}, respectively. 

Figure \ref{fig:heatmap_corruption} demonstrates a significant improvement in the robustness to all the natural corruptions when learning depth with transformers instead of convolutional networks.
Figure \ref{fig:heatmap_pgd} further shows a general improvement in adversarial robustness when learning either depth or ego-motion with transformers instead of convolutional networks.
Finally, Figure \ref{fig:bar_flip} shows an improvement in robustness to symmetrically flipped adversarial attacks when depth is learned using transformers instead of convolutional networks. Furthermore, depth estimation is most robust when both depth and ego-motion are learned using transformers. 

Therefore, MT-SfMLearner, where both depth and ego-motion and learned with transformers, provides the highest robustness and generalizability, in line with the studies on image classification~\cite{paul2021vision,bhojanapalli2021understanding}.
This can be attributed to their global receptive field, which allows for better adjustment to the localized 
deviations by accounting for the global context of the scene. 


\subsection{Intrinsics}
\label{subsec:intrinsics}
\begin{figure*}[t]
\centering
 \includegraphics[width=\textwidth]{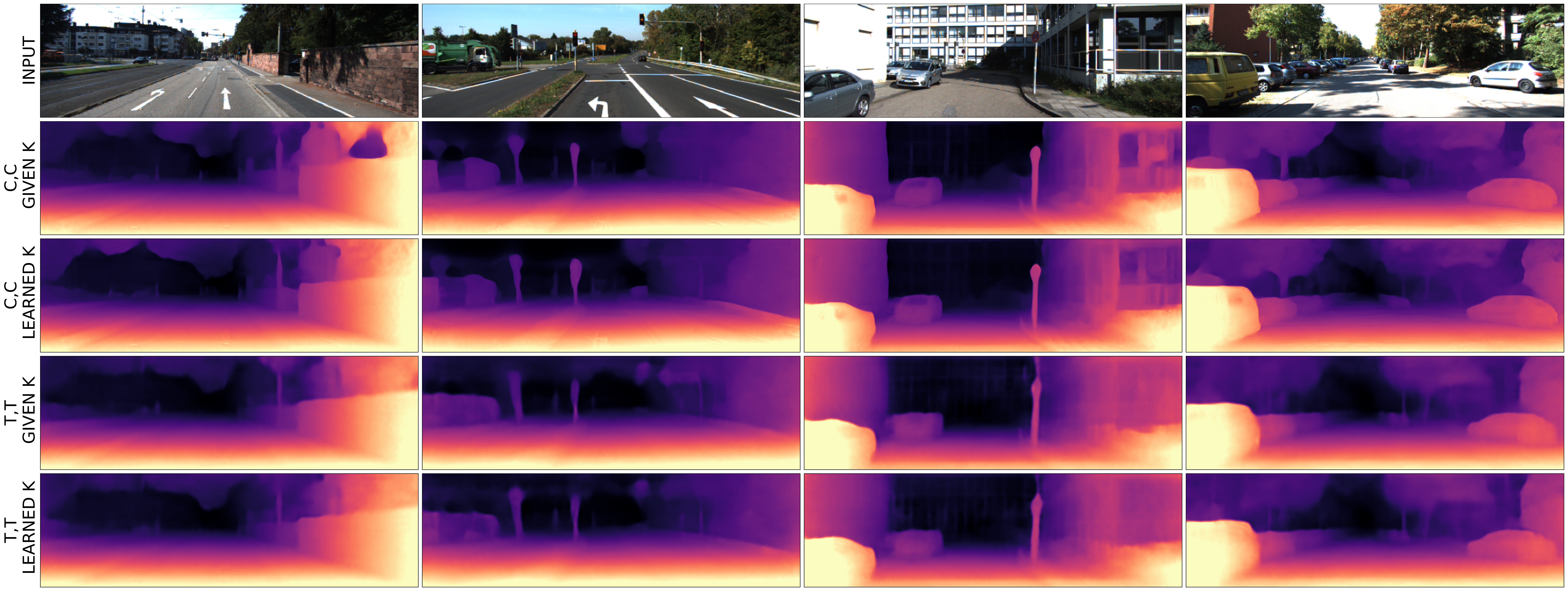}
  \caption{Disparity maps for qualitative comparison on KITTI, when trained with and without intrinsics (K). The second and fourth rows are same as the second and the fifth rows in Figure \ref{fig:ablation_vis}.} 
\label{fig:intrinsic_vis}
\end{figure*}

Here, we evaluate the accuracy of our proposed method for self-supervised learning of camera intrinsics and
its impact on the depth estimation performance. As shown in Table \ref{tab:intrinsics_table}, the percentage error for intrinsics estimation is low for both convolutional and transformer-based methods trained on an image size of $640\times192$. Moreover, the depth \textit{error} as well \textit{accuracy} metrics are both similar to when the ground truth intrinsics are known a priori. This is also observed in Figure \ref{fig:intrinsic_vis} where the learning of intrinsics causes no artifacts in depth estimation. 

\begin{table*}[h]
\centering
\caption{Percentage error for intrinsics prediction and impact on depth estimation for KITTI Eigen split.  
} 
\resizebox{\textwidth}{!}{
\begin{tabular}{|c|c|cccc|cccc|ccc|}
\hline
\multirow{2}{*}{\textbf{Network}} & \multirow{2}{*}{\textbf{Intrinsics}} & \multicolumn{4}{c|}{\textbf{\cellcolor{red!25} Intrinsics Error(\%) $\downarrow$}} &
\multicolumn{4}{c|}{\textbf{\cellcolor{red!25} Depth Error$\downarrow$}} &
\multicolumn{3}{c|}{\textbf{\cellcolor{blue!25}Depth Accuracy$\uparrow$}} \\ 
\cline{3-13} & & $f_{x}$ & $c_{x}$ & $f_{y}$ & $c_{y}$ & Abs Rel & Sq Rel & RMSE & RMSE log & $\delta<1.25$ & $\delta<1.25^2$ & $\delta<1.25^3$ \\ \hline \hline
\multirow{2}{*}{}
  \multirow{2}{*}{C,C} & Given & - & - & - & - & 0.111 & 0.897 & 4.865 & 0.193 & 0.881 & 0.959 & 0.980 \\
    & Learned & -1.889 & -2.332 & 2.400 & -9.372 & 0.113 & 0.881 & 4.829 & 0.193 & 0.879 & 0.960 & 0.981 \\
  \hline
  \multirow{2}{*}{T,T} & Given & - & - & - & - &  0.112 & 0.838 & 4.771 & 0.188 & 0.879 & 0.960 & 0.982 \\
    & Learned & -1.943 & -0.444 & 3.613 & -16.204 & 0.112 & 0.809 & 4.734 & 0.188 & 0.878 & 0.960 & 0.982\\
  \hline
\end{tabular}}
\label{tab:intrinsics_table}
\end{table*}

\begin{figure}[t]
\centering
  \includegraphics[width=\linewidth]{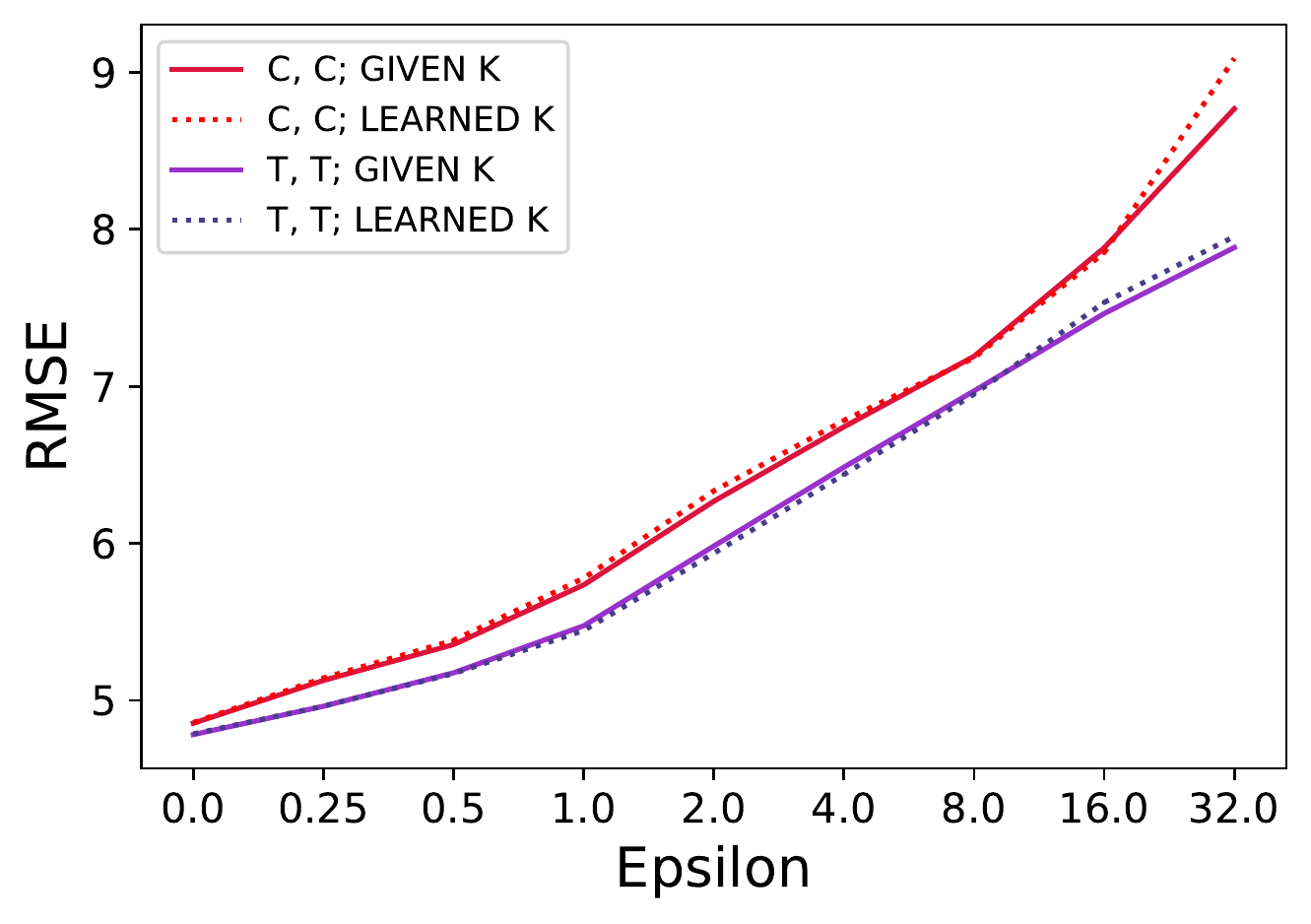}
  \caption{Mean RMSE for adversarial corruptions of KITTI generated using PGD, when trained with and without ground-truth intrinsics (K).} 
\label{fig:adversarial_intrinsics}
\end{figure}

We also evaluate the models trained with learned intrinsics on all 15 natural corruptions as well as on PGD and symmetrically flipped adversarial examples. We report the mean RMSE ($\mu$RMSE) across all corruptions in Table~\ref{tab:intrinsics_corr}. The RMSE for depth estimation on adversarial examples generated by PGD method for all strengths is shown in Figure~\ref{fig:adversarial_intrinsics}. The mean RMSE ($\mu$RMSE) across all attack strengths for horizontally flipped and vertically flipped adversarial examples is shown in Table~\ref{tab:adv_flip_intr}.

We observe that the robustness to natural corruptions and adversarial attacks is maintained by both architectures when the intrinsics are learned simultaneously. Furthermore, similar to the scenario with known ground truth intrinsics, MT-SfMLearner with learned intrinsics has higher robustness than its convolutional counterpart. 


\begin{table}[]
\centering
\caption{Mean RMSE ($\mu$RMSE) for natural corruptions of KITTI, when trained with and without ground-truth intrinsics.}
\resizebox{0.69\linewidth}{!}{
\begin{tabular}{|c|c|c|}
\hline
\textbf{Architecture}  & \textbf{Intrinsics} & \textbf{\cellcolor{red!25}$\mu$RMSE$\downarrow$}   \\ \hline \hline
\multirow{2}{*}{C, C}   & Given      & 7.683 \\
   & Learned    & 7.714 \\ \hline
\multirow{2}{*}{T, T} & Given      & 5.918 \\
 & Learned    & 5.939 \\ \hline
\end{tabular}}
\label{tab:intrinsics_corr}

\end{table}

\begin{table}[]
\centering
\caption{Mean RMSE ($\mu$RMSE) for horizontal (H) and vertical (V) adversarial flips of KITTI, when trained with and without ground-truth intrinsics.}
\resizebox{0.99\linewidth}{!}{
\begin{tabular}{|c|c|c|c|}
\hline
\textbf{Architecture}  & \textbf{Intrinsics} & \textbf{\cellcolor{red!25}$\mu$RMSE$\downarrow$ (H)} & \textbf{\cellcolor{red!25}$\mu$RMSE$\downarrow$ (V)}   \\ \hline \hline
    \multirow{2}{*}{C, C}   & Given      & 7.909 & 7.354 \\
  & Learned    & 7.641 & 7.196 \\ \hline
\multirow{2}{*}{T, T} & Given      & 7.386 & 6.795 \\
 & Learned    & 7.491 & 6.929 \\ \hline
\end{tabular}}
\label{tab:adv_flip_intr}
\end{table}


\subsection{Efficiency}
We further compare the networks on their computational and energy efficiency to examine their suitability for real-time applications. 

In Table \ref{tab:efficiency}, we report the mean inference speed in frames per second (fps) and the mean inference energy consumption in Joules per frame for depth and intrinsics estimation for both architectures. These metrics are computed over $10,000$ forward passes at a resolution of $640 \times 192$ on an NVidia GeForce RTX 2080 Ti. 

Both architectures run depth and intrinsics estimation in real-time with an inference speed $> 30$ fps. However, the transformer-based method consumes higher energy and is computationally more demanding than its convolutional counterpart.

\begin{table}[]
\centering
\caption{Inference Speed (frames per second) and Energy Consumption (Joules per frame) for depth and intrinsics estimation using CNN- and transformer-based architectures.}
\resizebox{0.84\linewidth}{!}{
\begin{tabular}{|c|c|c|c|}
\hline 
\textbf{Architecture}  & \textbf{Estimate} &  \textbf{\cellcolor{blue!25}Speed$\uparrow$} & \textbf{\cellcolor{red!25}Energy$\downarrow$}\\ \hline \hline
\multirow{2}{*}{C,C}  & Depth  & 84.132      & 3.206 \\ 
 & Intrinsics &  97.498      & 2.908 \\ \hline
\multirow{2}{*}{T,T}  & Depth   &  40.215      & 5.999 \\ 
 & Intrinsics & 60.190      & 4.021 \\ \hline
\end{tabular}}
\label{tab:efficiency}
\end{table}


\section{\uppercase{Conclusion}}

This work is the first to investigate the impact of transformer architecture on the SfM inspired self-supervised monocular depth estimation. Our transformer-based method MT-SfMLearner performs comparably against contemporary convolutional methods on the KITTI depth prediction benchmark. Our contrastive study additionally demonstrates that while CNNs provide local spatial bias, especially for thinner objects and around boundaries, transformers predict uniform and coherent depths, especially for larger objects due to their global receptive field. We observe that transformers in the depth network result in higher robustness to natural corruptions, and transformers in both depth, as well as the ego-motion network, result in higher robustness to adversarial attacks. With our proposed approach to self-supervised camera intrinsics estimation, we also demonstrate how the above conclusions hold even when the focal length and principal point are learned along with depth and ego-motion. 
However, transformers are computationally more demanding and have lower energy efficiency than their convolutional counterparts.
Thus, we contend that this work assists in evaluating the trade-off between performance, robustness, and efficiency of self-supervised monocular depth estimation for selecting the suitable architecture.

\bibliographystyle{apalike}
{\small
\bibliography{references}}


\end{document}